\documentclass[letterpaper, 10 pt, conference]{ieeeconf}  

\IEEEoverridecommandlockouts                              

\usepackage[english]{babel}
\usepackage{amsmath}
\usepackage{amssymb}
\usepackage{booktabs}
\usepackage{siunitx}
\usepackage[dvips]{graphicx}
\usepackage{multirow}
\usepackage{amsfonts}
\usepackage{enumerate}
\usepackage{tabularx}
\usepackage{algorithm,algorithmic}
\usepackage{bm}
\usepackage{adjustbox}
\usepackage{xcolor}
\usepackage{siunitx}
\usepackage{booktabs}
\usepackage{mdframed}
\usepackage{adjustbox}
\usepackage{authblk}
\usepackage{color}
\usepackage{xurl}
\usepackage{subcaption}
\usepackage{cite}
\makeatletter
\let\NAT@parse\undefined
\makeatother
\usepackage{hyperref}
\usepackage{relsize}
\usepackage{float}
\usepackage{pifont}

\title{\LARGE \bf
Wind-Aware Optimal Trajectory Planning for Efficient Gliding\\ of Fixed-Wing Aerial Systems
}

\author{}

\author{Luca Morando$^{1,2}$, Nishanth Bobbili$^2$, Luca Masci$^{1,2}$, and Giuseppe Loianno$^2$
\thanks{$^1$The authors are with New York University, NY 10012, USA. {\tt\footnotesize email: \{luca.morando, lm5175\}@nyu.edu}.}
\thanks{$^2$The author is with the University of California Berkeley,
Department of Electrical Engineering and Computer Sciences,
Berkeley, CA 94720, USA. {\tt\footnotesize email: \{nishanth.bobbili, loiannog\}@eecs.berkeley.edu}.}
\thanks{This work was supported by the DARPA Albatross Grant HR00112590173. Approved for Public Release, Distribution Unlimited.}
}

\begin{document}

\maketitle
\thispagestyle{empty}
\pagestyle{empty}

\begin{abstract}
Gliding offers small fixed-wing UAVs extended endurance and silent operation but requires accurate energy management, especially under wind disturbances and obstacle constraints. Traditional Total Energy Control Systems based controllers regulate the trade between potential and kinetic energy reactively, often requiring fine-tuning and trim-conditions knowledge. In this work, we shift the regulation to the planning level and present a nonlinear, multi-cost trajectory planner for small UAV gliders. The method generates $\mathcal{C}^3$ continuous trajectories based on Bernstein polynomials, mapped into control commands through differential flatness, and re-planned online to match experimentally derived sink polar curves. A simulated netto variometer is integrated into the optimization to estimate air mass motion, constraining the glide to energy-balanced states. Consecutive gliding trajectories are linked by cruising segments computed through trajectories initialized on Dubins path-based waypoints, enabling hybrid missions that combine powered and unpowered flight. The approach is validated in CFD simulations and real-world experiments with a fixed-wing platform, showing reliable stabilization of sink rate, airspeed, and glide ratio under wind gusts and in presence of obstacles.


\end{abstract}


\begin{figure*}[t!]
\centering
\includegraphics[width=\textwidth, trim=0cm 0cm 0cm 0cm, clip]{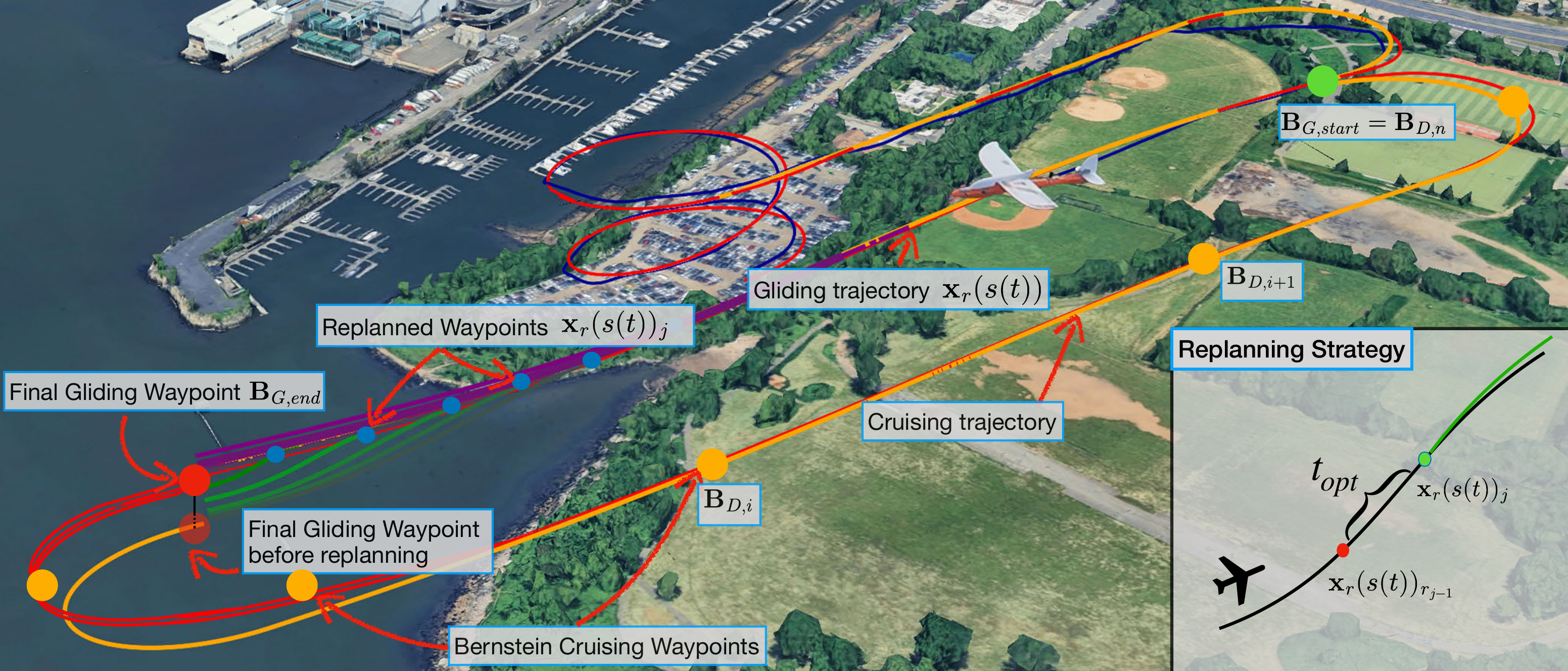}
\caption{Full-scale mission in Google Earth. The trajectory shows accurate replanning in both gliding and cruising states while the red waypoint offset highlights altitude saved through continuous adjustment in windy conditions.}
\label{fig:fig1}
\vspace{-20pt}
\end{figure*}

\section{Introduction}
Gliding, the art of unpowered flight, has fascinated humans since the earliest manned flights, inspiring advances in both mechanics and software while drawing lessons from nature. In soaring, long-range glides are essential to transition between lift sources through still air, and in manned aviation they are critical to maximize distance after sudden thrust loss \cite{vavna2018any, Deal24}. Gliding is also increasingly relevant for small Fixed-Wing Unmanned Aerial Vehicles (FW-UAVs), enabling tasks such as environmental monitoring \cite{GREEN2019465} and low-altitude surveillance \cite{Jaimes} with reduced motor noise and extended endurance. By exploiting natural lift sources like thermals \cite{renno1995quasi, Gall24} or ridge lift \cite{Lim24}, UAVs can reserve battery power for onboard systems and computing.

Aircraft without thrust entirely rely on gravitational potential energy, requiring a continuous trade-off between potential and kinetic energy to stay airborne. This trade-off must account for velocity limits, motion feasibility, and precise sensing of both aircraft state and aerodynamic effects, in order to maintain flight within the safe envelope \cite{VANDENBRANDT2018628} and avoid stall or structural damage at extreme speeds \cite{Deal24}. 

Small FW-UAVs face additional challenges: their lower Reynolds number reduces achievable $L/D$, and their light mass and slower speeds make them more sensitive to wind disturbances, demanding rapid compensation at both planning and control levels \cite{Hao24}. Most of the existing solutions regulate this energy balance through altitude control, typically using modified Total Energy Control System (TECS) controllers that adjust pitch to trade altitude for airspeed \cite{Xin24, Deal24, Oettershagen17}.
Traditional TECS controllers \cite{Ashwin22} respond reactively on sensor data, requiring careful tuning and knowledge of trim conditions, especially without thrust input. To overcome these limitations, we shift control to the planning level and propose a nonlinear, multi-cost trajectory planner for small fixed-wing gliders. Considering trajectories parametrized as Bernstein Polynomials~\cite{morando2025,kielas2019bebot}, the planner predicts optimal glide paths under wind and obstacle constraints, generating $\mathcal{C}^3$ continuous trajectories that are re-planned online to match the experimentally derived sink polar curve \cite{Oswaldo23} while respecting flight dynamics and controllability \cite{Tekles, Johannes}. Consecutive glide waypoints are connected via optimized cruising trajectories initialized from Dubins paths \cite{LugoCrdenas2014DubinsPG}, similar to \cite{bry2015aggressive}. The trajectory is mapped into control commands through differential flatness considering coordinated flight conditions.


To realistically optimize the generation of glide trajectories, we leverage the mathematical formulation of a simulated netto variometer, used to estimate the vertical speed of the surrounding air to detect stable glides or natural lift sources such as thermals or ridge effects\cite{tabor2018ardusoar, dupont1979accurate} and we directly incorporate it at the planning level in our trajectory optimization approach based on Bernstein Polynomials.

The main contributions of this paper are:
\begin{itemize}
\item A minimum-jerk, time-optimal, and wind-aware trajectory planner that generates glide paths that incorporate geometrical and dynamically constraints while respecting the path curvature and avoiding obstacles.
\item A variometer model represented with Bernstein polynomials, incorporated as a constraint in the trajectory optimization to predict sink rates accurately.
\item A continuous replanning strategy to overcome simplified stability assumptions for gliding based on the absence of wind.
\item Real-world experiments demonstrating effectiveness under wind and obstacle conditions, with performance compared to CFD simulations of the aircraft model.
\end{itemize}

\section{Related Works}
\label{Sec:Related_Works}
\vspace{-3pt}

Compared to VTOL rotorcraft, FW-UAVs offer longer endurance \cite{Autosoar18}, lower noise \cite{galles2020reducing}, and reduced battery use by exploiting aerodynamics \cite{Oettershagen19}, especially during wind-assisted glides. However, these benefits come with stronger coupled dynamics and greater planning complexity \cite{VANDENBRANDT2018628, Dobrokhodov2020}.
Safe glide trajectories remain underexplored despite their importance for emergency landings and soaring. Motion-primitive planners \cite{Atkins} and kinodynamic Rapidly Exploring Random Tree (RRT)* \cite{vavna2018any} ensure feasible descent paths but are computationally heavy for real-time use. More recent work \cite{Deal24} combines CL-RRT with a dissipative TECS controller for energy regulation and glide stablization, yet practical deployment is still limited.
Most existing methods rely on separate planners and energy-aware controllers, making velocity and altitude prediction along the path difficult. As noted in \cite{morando2025}, the field still lacks fast planners generating trajectories directly in the flat-output space, despite the efficiency of differential flatness in mapping planned states to control inputs \cite{Barry, Levin, Nieuwstadt}.
Building on gliding strategies from \cite{tabor2018ardusoar, Oettershagen17}, we propose a Bernstein-polynomial trajectory planner for both cruising and gliding flight. For actuated segments, we optimize a trajectory based on a sequence of waypoints sampled on a Dubins path \cite{bry2015aggressive}, while for gliding we generate Bernstein-based trajectories that predict and preserve energy balance, taking into account wind and obstacle avoidance. 
We incorporate the variometer model \cite{tabor2018ardusoar, Autosoar18} directly into the planner by representing it as a Bernstein polynomial and embedding it in the trajectory optimization. This predictive design yields energy-feasible glide paths, reducing the reliance on reactive control (e.g., TECS) to compensate for unplanned losses. In emergency landing scenarios, this ensures that feasible sites can be safely reached without the risk of stalling or undershooting. In soaring conditions, it enables the proactive use of favorable air masses to extend range and endurance beyond what control-level regulation alone can achieve.
As in \cite{hauser1997aggressive, Bry2015AggressiveFO, morando2025}, our trajectories are parameterized in space to maintain constant velocity. Unlike previous works such as \cite{Bry2015AggressiveFO, Liu}, we validate our approach on small FW-UAVs in challenging outdoor conditions and demonstrate real-time deployment,  bridging theory and practice.

\begin{figure}[t!]
\includegraphics[width=0.9\columnwidth, trim={1cm 0.5cm 3cm 0.5cm}, keepaspectratio]{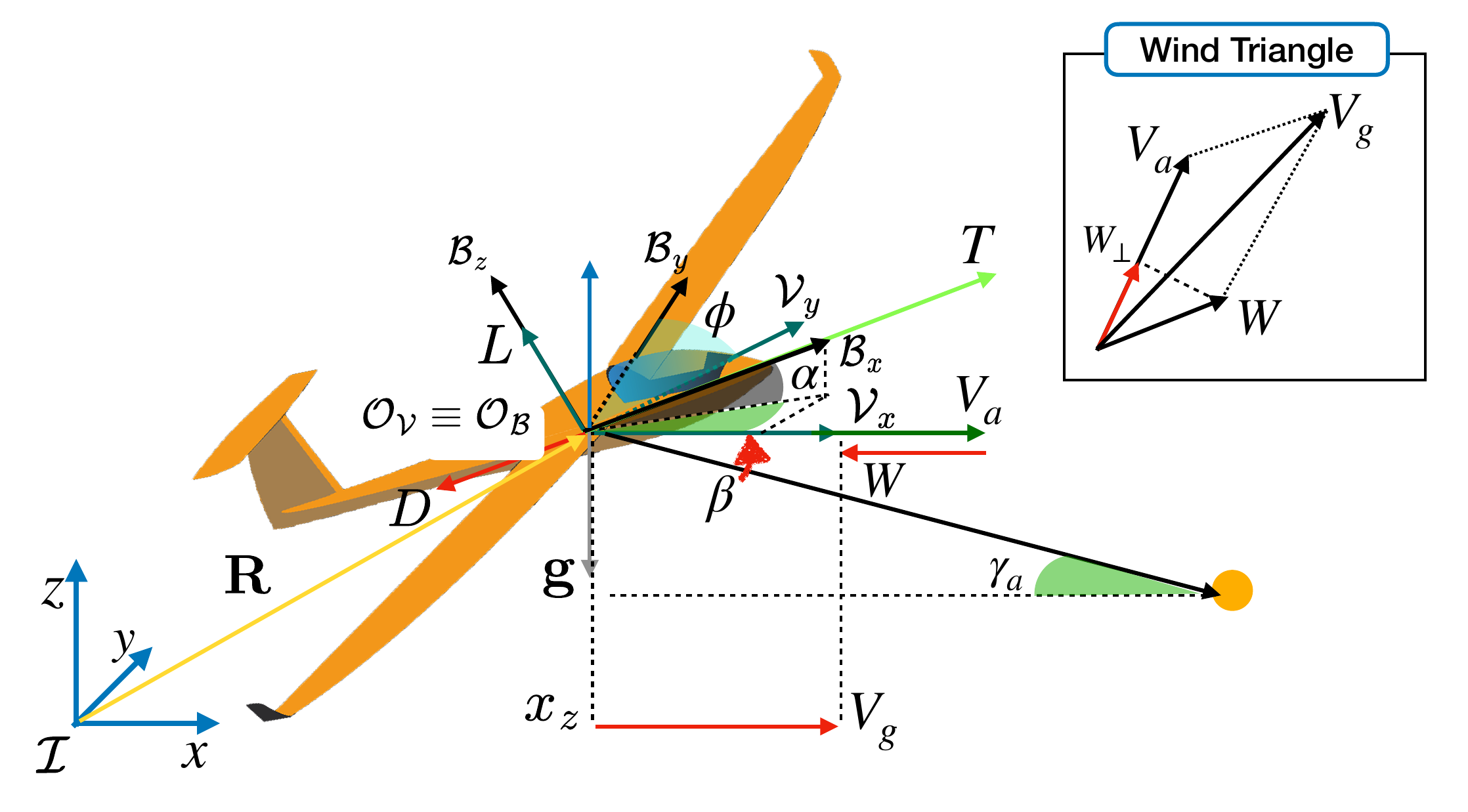}
  \caption{System overview and notation. The Body  $\mathcal{B}$ and Velocity $\mathcal{V}$ frames. The angle $\gamma_a$ defines the glide slope, which changes based on the wind $\mathbf{W}$ effects on the ground velocity $\mathbf{V}_g$, according to the wind triangle convention. 
}
 \label{fig:fig2}
\vspace{-20pt}
\end{figure}

\begin{figure*}[!t]
  \centering
  \includegraphics[width=\textwidth]{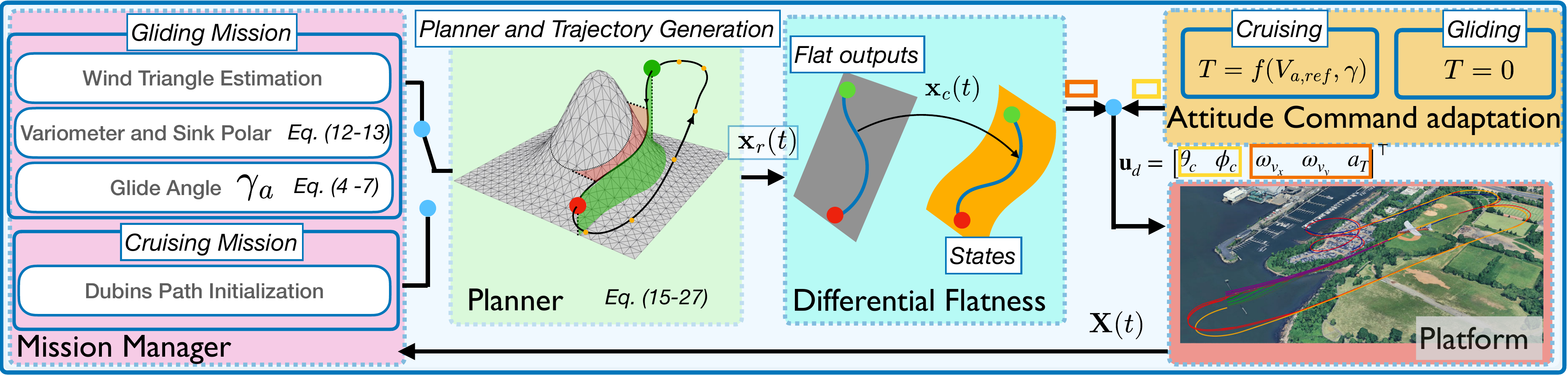}
  \caption{An overview of the proposed planning architecture. The trajectory is generated based on the aircraft’s energy state, flight mode, environment, and obstacles. The Differential flatness block maps the trajectory into attitude commands. 
  \label{fig:software_architecture}}
  \vspace{-20pt}
\end{figure*}

\section{System Modeling}
\label{sec:System_modeling}
As shown in Fig.~\ref{fig:fig2}, we adopt the following frame convention where the inertial frame is denoted by $\mathcal{I}$ while $\mathcal{B}$ denotes the vehicle’s rigid body frame that it is aligned with the FW's longitudinal, lateral, and vertical axes. To consider the aerodynamics effect, we also define the velocity frame $\mathcal{V}$ which is centered at $\mathcal{B}$ and is rotated with respect to the it by the sideslip angle $\beta$ and the angle of attack $\alpha$, therefore always keeping the corresponding $\mathcal{V}_x$ axis aligned with the direction of the aircraft velocity.
The vehicle system dynamics, defined in the inertial frame $\mathcal{I}$ as
\begin{equation}
    \begin{split}
    \mathbf{\dot{x}} = V\mathbf{R}\mathbf{e}_1,&~\mathbf{\ddot{x}} = \mathbf{g} + \mathbf{R}\mathbf{a}_v,\\
    \mathbf{\dot{R}} = \mathbf{R}\hat{\bm\omega}_v,~\dot{\bm{\omega}}_v &= \mathbf{J}^{-1}(-\bm{\omega}_v \times \mathbf{J}\bm{\omega}_v + \bm{\tau}),\label{eq:system_dynamics4}
\end{split}
\end{equation}
where the FW state is $\mathbf{X} = \{\mathbf{x}, \mathbf{\dot{x}}, \mathbf{R}, \bm{\omega}_v\}$, where $\mathbf{x} \in \mathbb{R}^3$ with 
$\mathbf{x} \in \mathbb{R}^3$ represents the position of FW in $\mathcal{I}$,  
The rotation between $\mathcal{B}$ and  $\mathcal{V}$, is represented by $R_{\mathcal{V}}^{\mathcal{B}}(\alpha, \beta)$, 
while $\mathbf{R} \in SO(3)$ with $\mathbf{R}= \mathbf{R}_{\mathcal{B}}^{\mathcal{I}}\mathbf{R}_{\mathcal{V}}^{\mathcal{B}}$ denotes the rotation of $\mathcal{V}$ with respect to $\mathcal{I}$.
The velocity of the FW is represented by the velocity vector $\mathbf{\dot{x}} = [\dot{x}_{x}~\dot{x}_{y}~\dot{x}_{z}]^\top$, with a zero lateral velocity component ($\dot{x}_{y} = 0$) to satisfy the coordinated flight condition.
Additionally, $V = \lVert \dot{\mathbf{x}} \rVert$, $\mathbf{g} = [0~0~-g]^{\top}$ is the gravity vector in $\mathcal{I}$, $\mathbf{e}_1 = [1~0~0]^{\top}$ is the unit vector aligned with the $x$ direction of $\mathcal{V}$, $\mathbf{a_v} = [a_{v_x}~0~a_{v_z}]^\top$ contains the axial and normal accelerations, while $\boldsymbol{\omega}_v = [\omega_{v_x}~\omega_{v_y}~ \omega_{v_z}]^\top$ represents the angular velocity of $\mathcal{V}$ wrt. $\mathcal{I}$, expressed in $\mathcal{V}$ and $\hat{\bm\omega}_v$ its corresponding skew-symmetric matrix. Finally, $\mathbf{J} \in \mathbb{R}^{3 \times 3}$ describes the inertia acting on each direction of the body frame $\mathcal{B}$, while $\bm{\tau}$ expresses the torque applied on the system due to the action of control surfaces, like ailerons, elevators, and rudder. Additional details on the FW dynamics for coordinated flight conditions is available in \cite{morando2025} and \cite{hauser1997aggressive}.
As shown in Fig.~\ref{fig:fig2}, under powered flight, thrust $T$ acts along the longitudinal axis, balancing drag and wind-induced resistance. Once expressed in the inertial frame $\mathcal{I}$, we can define a direct relationship between the ground velocity vector $\mathbf{V}_g^{\mathcal{I}}$, the airspeed vector $\mathbf{V}_a^{\mathcal{I}}$, and the estimated wind vector $\mathbf{W}^{\mathcal{I}}$. 
Specifically, the wind vector $\mathbf{W}^{\mathcal{I}}$ modifies the ground velocity $\mathbf{V}_g^{\mathcal{I}}$ such that the airspeed is related to $\mathbf{V}_g^{\mathcal{I}}$ through the classical wind triangle:
$\mathbf{V}_a^{\mathcal{I}} = \mathbf{V}_g^{\mathcal{I}} - \mathbf{W}^{\mathcal{I}}$.

\subsection{Aerodynamics for Longitudinal Gliding}
\label{sec:Aerodynamics_gliding}

The aerodynamic forces on the airframe decompose into lift $L$, drag $D$, and weight force $F_g$ 
\begin{equation}
    \begin{split}
    L &= \frac{1}{2}\rho V_a^2 S C_L(\alpha, \beta)  ,\\ 
    D &=\frac{1}{2}\rho V_a^2 S (C_{D0} + kC_L(\alpha)^2 + k_{\beta} \beta^2), ~
    F_g = mg, 
    \label{eq:lift_drag}
\end{split}
\end{equation}
where $\rho$ is the air density, $V_a=\lVert \mathbf{V}_a^{\mathcal{I}} \rVert$ the airspeed, $S$ the wing area, $C_L(\alpha, \beta)\approx C_L(\alpha)$ the Lift Coefficient, $k_\beta$ accounts for induced drag from sideslip and $m$ is the mass.  For simplicity, we consider
\begin{equation}
    C_L(\alpha) = \pi AR e,~kC_L(\alpha)^2 + k_{\beta}\beta^2 = \frac{C_L(\alpha)^2}{\pi ARe},
    \label{eq:Cl_semplification}
\end{equation}
with $AR$ the aspect ratio and $e$ the Oswald efficiency factor.



In steady-state gliding flight with no wind, lift $L$ and drag $D$ balance the gravity force $F_g$ such that no horizontal or vertical acceleration acts on the airframe. This defines a preservation in the exchange of kinetic to potential energy, validating the principle of energy conservation.  Considering that $V_a$  and the glide angle $\gamma_a$ (see Fig.~\ref{fig:fig2}) are constant, lift and drag are
\begin{equation}
L = mg\cos\gamma_a,~ \quad D = -mg\sin\gamma_a,
\label{eq:lift_drag_gamma}
\end{equation}
where $\gamma_a$ is the glide angle. Substituting the drag defined in eq.~(\ref{eq:lift_drag}) in eq.~(\ref{eq:lift_drag_gamma})  using $C_L(\alpha)$ in eq.~(\ref{eq:Cl_semplification}), we obtain 
\begin{equation}
\frac{1}{2}\rho V_a^2 S \left(C_{D0} + \frac{C_L^2(\alpha)}{\pi ARe}\right) = -mg\sin\gamma_a.
\label{eq:lift_drag_gamma_new}
\end{equation}

With further analysis on Fig.~\ref{fig:fig2}, we can consider the lift coefficient directly affected by $\gamma_a$, such that $C_L(\alpha) = mg\cos\gamma_a$. 
Considering that  $(1-\sin^2\gamma_a) = \cos^2\gamma_a$, we can plug this value of $C_L(\alpha)$ directly in eq.~(\ref{eq:lift_drag_gamma_new}) and obtain
\begin{equation}
    mg \sin^2 \gamma_a - \left(\frac{1}{2} \rho V_a^2 S \pi ARe\right) \sin\gamma_a - \eta = 0,
    \label{eq:lift_drag_gamma_2}
\end{equation}
where $\eta = ((0.5 \rho V_a^2 S)^2\pi AReC_{D0} + (mg)^2)/mg$. 

The quadratic equation can be solved for $sin \gamma_a$ showing that for each airspeed $V_a$ there exists a unique relative glide angle that yields steady-state flight without acceleration
\begin{equation}
    \sin \gamma_a = \frac{\frac{1}{2} \rho V_a^2 S \pi ARe - \upsilon}{2mg},
    \label{eq:lift_drag_gamma_3}
\end{equation}
where $\upsilon = \sqrt{(\frac{1}{2} \rho V_a^2 S)^2 \pi AR (\pi AR e + 4 C_{D0} ) + 4 m^2g^2}$.

Considering eq.~(\ref{eq:lift_drag_gamma_3}), for any airspeed $V_a$ there exists a unique glide angle $\gamma_a$ which allows a stable glide. However, given the relationship between lift and drag, we can assume that there exists an optimal gliding airspeed, denoted as  $V_a^*$ that minimizes the glide angle $\gamma_a$, therefore maximizing the lift-to-drag ratio $(L/D)$. 

Following \cite{lawrance2011autonomous}, by substituting eq.~(\ref{eq:lift_drag_gamma_3}) into the drag expression in eq.~(\ref{eq:lift_drag_gamma}), we obtain the drag $D$ as a function of $V_a$. The minimum-drag airspeed  $V_a^*$ corresponds to a saddle point of this function, which can be determined by differentiating with respect to $V_a$. Solving the derivative for its roots yields the stationary points of the curve, one of which corresponds to the minimum-drag airspeed
\begin{equation}
V_{a}^* = \sqrt[4]{(4m^2g^2)/ \rho^2 S^2 C_{D0}(\pi AR e + 4 C_{D0})}.
\end{equation}
At this airspeed, we obtain the maximum $(L/D)$ ratio
\begin{equation}
\max(L/D) = -\frac{1}{\tan \gamma_a} \\
          = \frac{\sqrt{1 - \sin^2 \gamma_a}}{-\sin \gamma_a} \\
          = 0.5 \sqrt{\frac{k}{C_{D0}}}.
\end{equation}

as similarly obtained in \cite{Owen2015,lawrance2011autonomous}.
\subsection{Variometer and Total Specific Energy}
\label{sec:Vario_and_energy}
In establishing the glide angle providing the maximum drag-to-lift ratio, the key assumption is that the surrounding air is still, neglecting wind disturbances or vertical lift from thermals \cite{Autosoar18}. To compensate for these effects and maintain a constant airspeed $V_a$ thereby ensuring a stable glide, we rely on variometer data. As introduced in \cite{tabor2018ardusoar}, a variometer estimates the vertical speed of the surrounding air, allowing the planner to detect sources of lift, such as thermals, and adjust the trajectory accordingly. Although no sensor directly measures vertical airspeed, the variometer infers it from the total specific energy capture changes of the aircraft due to lift or energy trade between altitude and speed. Following \cite{tabor2018ardusoar}, the rate of exchange of energy is represented by the first derivative of the total energy of the aircraft during glide. Considering the optimal airspeed $V_a^*$ derived in the previous section, we can write the energy equation as
\begin{equation}
    E = mg x_z + \frac{1}{2}m {V_a^*}^{2},
\end{equation}
Dividing by $mg$ and computing the first derivative, we obtain the energy rate as
\begin{equation}
    \frac{dE}{dt} = \dot{E}  = \dot{x}_z + \frac{V_a^* \dot{V}_a^*}{g}.
\end{equation}
The terms in these equations are obtained from onboard estimators or directly from the aircraft sensors. In particular, $\dot{x}_z$ is the vertical velocity in the inertial frame $I$, as in eq.~(\ref{eq:system_dynamics4}), and $\dot{V}_a^* $ is the airspeed time derivative, obtained using a Savitzky golay filter \cite{gallagher2020savitzky}, similarly to~\cite{Autosoar18}.

To estimate the true vertical speed of the surrounding air, $\dot{E}$ 
 is corrected with the aircraft’s natural sink rate
 \begin{equation}
    V_z(V_a, \phi) = V_a^* \left(\frac{P}{C_L} +\frac{B C_L}{\cos^2 \phi}\right),
    \label{eq:vario}
    \end{equation}
where $P$ and $B$ are airframe-specific regression constants that can be obtained fitting them from experimental flight data (see Section~\ref{sec:Vario_and_energy}).
 The netto variometer signal is
\begin{equation}
\dot{E}_{net} = \dot{E} +  V_z(V_a^*, \phi) = \dot{x}_z + \frac{V_a^* \dot{V}_a^*}{g} + V_z(V_a^*, \phi).
\end{equation}
 This corrected measure $\dot{E}_{net}$, called the netto variometer, reads $0~\si{m/s}$ when the aircraft glides at its nominal sink rate in still air relative to the gliding airspeed $V_a^*$. The sink rate $V_z(V_a^*, \phi)$ can be obtained from the glider drag polar curve, experimentally or via Computational Fluid Dynamics (CFD).

\subsection{Differential Flatness and Trajectory Parameterization}
\label{sec:differential_flat}

In this section, we revisit the differential flatness formulation for fixed-wing systems, parameterized by airspeed $V_a$ and traveled distance $s$ (see Fig.~\ref{fig:software_architecture}), to show how planning maps to control. A system is flat if its states and inputs can be expressed via flat outputs and their derivatives. Following~\cite{morando2025}, 
the trajectory tracking problem considering the state $\mathbf{x}(t)$ is solved using a PID loop
\begin{equation}
    \mathbf{x}_{c}^{(3)} = \mathbf{x}_{r}^{(3)} + k_2 \ddot{\mathbf{e}} + k_1 \dot{\mathbf{e}} + k_0 \mathbf{e},
    \label{eq:pid}
\end{equation}  
where $\mathbf{e} = \mathbf{x}_r - \mathbf{x}$ and $\mathbf{x}_{c}^{(3)}$ is the third derivative input to the flat model, from which it is possible to obtain the desired geometric attitude of the aircraft expressed as rotation matrix $\mathbf{R}_c$. We can then derive, the input commands  
$\mathbf{u}_{c} = [\theta_c~\phi_c~\omega_{v_x}~\omega_{v_y}~T]^\top$,  
with $\theta_c,\phi_c,\psi_c$ defining the Euler angles attitude commands and $T$ the thrust.
To avoid unsafe thrust variations from abrupt velocity or acceleration changes, the trajectory is parameterized by path length $s(t)$ at reference airspeed $V_{a,ref}$ \cite{morando2025, hauser1997aggressive}, yielding $\mathbf{x}_{r}=\mathbf{x}_{r}(s(t))$.


\section{Trajectory Planning}
\label{sec:planner}

In this section, we address the design of constrained, dynamically feasible, and energy-aware trajectories by formulating a nonlinear optimization problem that exploits the differential flatness and represents trajectories using Bernstein polynomials. These show interesting properties in terms of smoothness and imposing global spatial constraints compared to time-based polynomials~\cite{kielas2019bebot,kielas2022bernstein}. For a given trajectory $m_j$ belonging to a set $M$ such that $j \in (0, M)$, this is represented by the following structure of degree $n$
\begin{equation}
C_{n,m_j}(t) = \sum_{i=0}^{n}\mathbf{p}_{i,n}^{m_j}\beta^n_i(t),   \quad t\in [t_0, t_f],
\label{eq:bernstein_equation}
\end{equation}
where $\mathbf{p}_{i,n}^{m_j}$ are the Bernstein coefficient or control points of  size $n$ control, and $\beta^n_i(t)$ is the Bernstein basis. The $k^{\text{th}}$ derivative of the polynomial can be obtained as
\begin{equation}
\frac{d^{k}}{dt^{k}}C_{n,m_j}(t) = \frac{n!}{(n-k)!(t_f - t_0)^k} \sum_{i=0}^{n-k} {{\mathbf{p}}^{{m_j}^{'}}_{i,n-k}}\beta_{i}^{n-k}(t),
\label{eq:bernstein_derivative}
\end{equation}
with ${\mathbf{p}}^{{m_j}^{'}}_{i,n-k} = \mathbf{p}_{i,n}^{m_j}\mathbf{D}_k$ and $\mathbf{D}_k = \text{diag}(\mathbf{c}\circledast^k, \mathbf{c}\circledast^k, \cdots, \mathbf{c}\circledast^k)$ is the Differential matrix with $ \mathbf{c} = [-1, 1]$ convoluted $k$ times. Considering $M+1$ waypoints, a full trajectory $\mathbf{x}_{r}(t)$ can be modeled by stacking together $M$ Bernstein polynomials connected at the extremal points as





\begin{equation}
\mathbf{x}_{r}(t)  = 
\begin{cases} 
    \sum_{i=0}^{n}\mathbf{p}_{i,n}^{m_1}\beta_i^n(T_1 - t) \quad \text{for} \ t\in [0, T_1]\\
    \sum_{i=0}^{n}\mathbf{p}_{i,n}^{m_2}\beta_i^n(T_2 - t) \quad \text{for} \ t\in [T_1, T_2] \\
    \vdots \\ 
    \sum_{i=0}^{n}\mathbf{p}_{i,n}^{M}\beta_i^n(T_{M}-t) \quad \text{for} \ t\in [T_{M-1}, T_M]
\end{cases}
\label{eq:piecewise_bernstein_equation}
\end{equation}
where $\mathbf{p}_{i,n}^{m_j}$ is the $i^{th}$ control point of the $m_j$ sub trajectory, with $j \in [1, M]$, and the time instants $T_1, T_2, \dots, T_M$ represent the allocated time for each of sub trajectory. 

Building on this framework, we formulate the trajectory design as a nonlinear optimization problem
\begin{equation}
\begin{split}
\text{min} \quad & \sigma_0 J + \sigma_1 T  +  \sigma_2 W \\
\text{s.t.} \quad & \mathbf{y} < \mathbf{G(t)} \mathbf{p}_d < \mathbf{z}
\label{eq:optimization_general}
\end{split}
\end{equation}
where
\begin{itemize}
    \item $J = \int_0^{t_{f}} \left\|\mathbf{x}_{r}^{(3)}(t)\right\|^2 \, dt$ 
is the integral of the squared trajectory jerk.
\item $T = \sum_{j = 0}^f t_j$ is the total traversal time of the polynomial segments $j \in [0,t_f]$.
\item $W_{\perp} = - \sum_{j = 0}^n \mathbf{W}_p $ accounts for wind effects at each derived control point of the polynomial, maximizing $\mathbf{W}_p= \mathbf{W}^I \cdot \dot{\mathbf{x}}_{r}(t)$  defined as the projection of the wind vector $\mathbf{W}^I$ along the trajectory velocities $\dot{\mathbf{x}}_{r}(t)$, described through the derivative $\dot{\mathbf{p}}_d$ of the $n$ optimization coefficients for the $XY$ directions. 
%
\end{itemize}
Three optimization weights $\sigma_0$, $ \sigma_1$, and $ \sigma_2$ allow trading off between minimum jerk, minimum traversal time, and wind exploitation.
The vector $\mathbf{p}_d$ contains the polynomial coefficients, and $\mathbf{G}(t_j)$ is the constraint matrix evaluated at segment of duration $t_j$. 
The lower and upper bounds of the constraints are $\mathbf{y}$ and $\mathbf{z}$ respectively, are defined to  impose the following conditions
 \begin{itemize}
    \item Respect the geometry for coordinated flight conditions.
    \item Bounding the maximum curvature $\kappa$ of the trajectory.
    \item Bounding the sink rate $V_z$ so that the measurements from the netto variometer satisfy  $\dot{E}_{net} = 0$. 
    \item Continuous adaptation to local wind estimation and presence of obstacles
\end{itemize}
Constraints are enforced using the convex hull property of Bernstein polynomials, which guarantees that the trajectory remains within the prescribed bounds

\begin{enumerate}
\renewcommand{\labelenumi}{\arabic{enumi}.}

    \item \textit{Endpoint constraint.}
    Considering a starting time $t_0$ and an ending time $t_{f}$, we constrain $\mathbf{x}_{r}$ at the reference waypoints position $\mathbf{x}_{r}$, velocity $\dot{\mathbf{x}}_{r}$, and acceleration $\ddot{\mathbf{x}}_{r}$
     \begin{equation}\begin{aligned}
        C_{n,0}^{(k)}(t_0) = \mathbf{x}^{(k)}(t_0), \qquad C_{n,M}^{(k)}(t_f) = \mathbf{x}^{(k)}(t_f) 
    \end{aligned}.\end{equation}

    \item \textit{Continuity Constraints.}
    The goal is to ensure the continuity in position and higher derivatives of the trajectory $\mathbf{x}_r(t)$ at the junction of the $M$ sub trajectories as


    \begin{equation}\begin{aligned}
        C_{n,m}(t_{f}) = C_{n,m+1}(t_{{0}}).
    \end{aligned}\end{equation}

     \item \textit{Curvature or angular rate constraint.}
     The curvature of a trajectory, defined as $\kappa = f(\dot{\mathbf{x}}_{r_x}, \dot{\mathbf{x}}_{r_y}, \ddot{\mathbf{x}}_{r_x}, \ddot{\mathbf{x}}_{r_y})$ evaluated from $t_0$ to $t_f$ and differently from \cite{morando2025}, can be directly expressed as a bernstein polynomial given its natural relationship with the derivative of the heading angle of the trajectory $\Psi(t) = tan^{-1}(\frac{\dot{\mathbf{x}}_{r_y}}{\dot{\mathbf{x}}_{r_x}})$ as
    \begin{equation}\begin{aligned}
        \dot{\Psi}(t) = \omega(t) =  \frac{\dot{\mathbf{x}}_{r_x} \ddot{\mathbf{x}}_{r_y} - \ddot{\mathbf{x}}_{r_x}\dot{\mathbf{x}}_{r_y} }{\dot{\mathbf{x}}_{r_x}^2 + \dot{\mathbf{x}}_{r_y}^2}.
        \label{eq:curvature}
    \end{aligned}\end{equation}
    
     \item \textit{Ground Cruising Speed Magnitude.}
     Given the presence of wind, we model the magnitude of the trajectories velocities to respect the wind triangle constraint
     expressed in Section \ref{sec:Aerodynamics_gliding}. Given the squared norm of $\dot{\mathbf{x}}_{r}$ as 
         \begin{equation}\begin{aligned}
          \| \dot{\mathbf{x}}_{r} \|^2 = \mathbf{V}_g =  \dot{\mathbf{x}}_{r_x}\dot{\mathbf{x}}_{r_x} + \dot{\mathbf{x}}_{r_y} \dot{\mathbf{x}}_{r_y},
          \label{eq:Vg_bern}
    \end{aligned}\end{equation}

    we can correct the nominal cruising speed given the wind magnitude $\mathbf{W}$ imposing the following wind triangle constraint
     \begin{equation}\begin{aligned}
      \mathbf{V}_g - \mathbf{W} - \xi \leq \| \dot{\mathbf{x}}_{r} \|^2  \leq  \mathbf{V}_g   - \mathbf{W} + \xi,
      \end{aligned}\end{equation}

      where $\xi$ is defined as a small constraint relaxation value.
      
      \item \textit{Sink Rate Constraint.}
    To ensure that the gliding trajectory respects the aircraft’s nominal sink rate $V_{z, nom}(V_a, \phi)$, and maintains an energy-balanced state ($\dot{E}_{net} = 0$ and $V_a = V_{a, ref}$, we integrate the energy estimator, represented in Fig.~\ref{fig:software_architecture} and described in Section~\ref{sec:Vario_and_energy}, directly as a constraint as 
     \begin{equation}\begin{aligned}
     \mathbf{V}_{z, nom}(\mathbf{V}_a, \mathbf{\phi}) - \xi \leq \dot{\mathbf{x}}_{r_x} \leq   \mathbf{V}_{z, nom}(\mathbf{V}_a, \mathbf{\phi}) + \xi,
     \end{aligned}
     \end{equation}
     Similarly, the nominal sink rate $ \mathbf{V}_{z, nom}(\mathbf{V}_a, \phi)$ can be expressed in terms of a Bernstein Polynomial with $n$ control points, as for eq.~(\ref{eq:vario})
     \begin{equation}\begin{aligned}
     \mathbf{V}_{z, nom}(V_a, \phi) = \mathbf{V}_a\left(\frac{P}{C_L} + B C_L (1 + \tan{\phi}^2)\right),
     \end{aligned}\end{equation}

    where $\tan{\phi} = - \frac{\mathbf{V}_a^2 * \kappa}{g}$ defines an approximation to express the squared cosine as Bernstein polynomial, where $\kappa$ is polynomial describing the curvature as defined above in eq.~(\ref{eq:curvature}).
    
    \item \textit{Obstacle and No Fly Zone Avoidance.}
   As shown in Fig.~\ref{fig:software_architecture}, a Gaussian-shaped obstacle is defined as $\mathbf{O} = [x_o, y_o, h, \sigma_x, \sigma_y]$ with center $(x_o, y_o)$, height $h$ and variance $(\sigma_x, \sigma_y)$.
    The obstacle is approximated by a Bernstein polynomial $C_{n, obs} $ whose position is defined as $\mathbf{x}_{obs}$, allowing the Convex Hull property to enforce a safe distance from the trajectory.
    The squared distance in the XY plane is
     \begin{equation}\begin{aligned}
     \mathbf{d}^2(t) = (\mathbf{x}_{obs,x} - \mathbf{x}_{x})^2 + (\mathbf{x}_{obs,y} - \mathbf{x}_{y}) ^2
     \end{aligned},\end{equation}
     \begin{equation}
      d^2_{safe} \leq \mathbf{d}^2(t) \leq \infty
.\end{equation}
Since the required obstacle representation diescretization due by the coefficients position $C_{n, obs}$, a 2D formulation suffices to maintain safety, preventing the trajectory from passing above or inside the gap between obstacle coefficients. 
    
\end{enumerate}

In order to continuously compensate for updated wind estimations along the course of the gliding treajectory, we design a replanning technique visible in Fig.~\ref{fig:fig1}. The trajectory is continuously updated $\mathbf{x}(s(t))_{r, j}$ in order to keep the aircraft stable, also adjusting the terminal part $\mathbf{x}(s(t_f))_{r, j}$ so that $V_a = V_{a, ref}$ and $\dot{E} = 0$.

\section{Results}
\label{sec:Experimental_Results}

\begin{figure}[!t]
  \centering
    \includegraphics[width=\columnwidth]{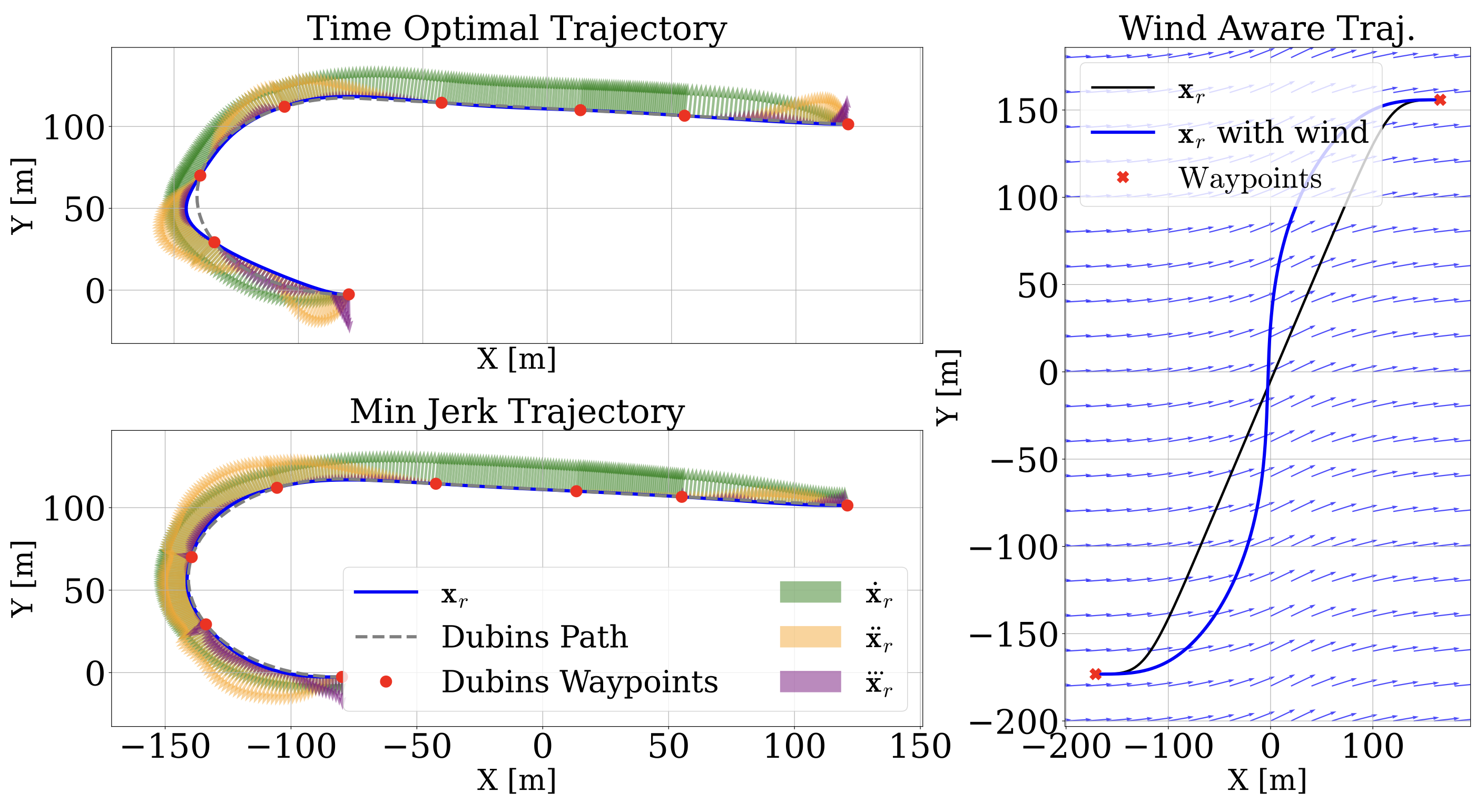}
    \caption{Trajectory optimization with more weight on a) time cost $\sigma_1$, b) jerk cost $\sigma_0$ and c) wind cost $\sigma_2$. }
    \label{fig:traj_opt}
  \vspace{-20pt} 
\end{figure}

In this section, we outline the mission planning based on continuous optimization of Bernstein Polynomials as anticipated in Fig. \ref{fig:fig1}, using the CASADI tool running the IPOPT solver~\cite{WachterBiegler2006}.
Our modular framework, visualized in Fig. \ref{fig:fig2}, is validated through simulations and real world experiments in a large outdoor field. The FW aircraft used for our experiments and shown in Fig.~\ref{fig:fig1}, is a Strix Stratosurfer, equipped with an OrangeCube Flight Controller running PX4 for low-level attitude control. Onboard computation is handled by an NVIDIA\textsuperscript{{\textregistered}} Jetson Xavier Orin board, running Ubuntu 22.04 and the ROS2\footnote{\url{www.ros.org}} framework for intra-processes communications. Localization is based on a Drotek\textsuperscript{{\textregistered}} F9P GNSS receiver integrated with the PX4 EKF2-based state estimator, while airspeed is measured via CAN based sensor.
The planner and trajectory manager, as illustrated in Fig. \ref{fig:software_architecture}, operate at $100~\si{Hz}$ to ensure smooth and continuous control. Mission data is also transmitted to the trajectory manager via the QGroundControl\textsuperscript{{\textregistered}} interface.

\subsection{Mission Planning}
For both simulation and real world experiments, missions are planned via the QGroundControl (QGC) interface as (see Fig. \ref{fig:fig1}). Waypoints are uploaded to the onboard computer using map coordinates, and each waypoint defines an actuated or gliding flight segment through cost function weights $\sigma_{0,2}$

\begin{itemize}
\item \textbf{Cruising Mode}: The aircraft performs actuated cruising to recover altitude lost during gliding. Minimum-jerk Bernstein trajectories are generated using a seed based on a set of waypoints $\mathbf{B}_D = [\mathbf{B}_{D, 0}, \dots, \mathbf{B}_{D, n}]  \in D$ upsampled from a Dubins path, represented as yellow spheres in Fig. \ref{fig:fig1}. 

\item \textbf{Gliding Mode}: The planner generates trajectories considering $T = 0$ between the waypoints $\mathbf{w}_G = [\mathbf{B}_{G, start}, \mathbf{B}_{G, end}] $ as visible in Fig.~\ref{fig:fig1}, activating all the cost objectives and constraints described in Section~\ref{sec:planner} including the gliding constraint and the wind cost term $\sigma_2$. The gliding segments are in yellow in Fig. \ref{fig:fig1}.  
\end{itemize}

For Cruising Mode, the weights are set as $\sigma_0 = 10$, $\sigma_1 = 0.1 $, $\sigma_2 = 0.1 $. This provide the planner more freedom in generating minimum jerk trajectories, instead of minimum time, creating slower but more round trajectories between the waypoints, relaxing as well the curvature constraint and reducing the optimization time. In Fig. \ref{fig:traj_opt} illustrates an optimized Bernstein trajectory, comparing the value of $J$ obtained in case of minimum time  and minimum Jerk optimization, respectively resulting in $J = 0.279~\si{m/s^3}$ and $J = 0.052~\si{m/s^3}$. The wind exploitation effect on the optimization is shown in Fig. \ref{fig:traj_opt} (right) with $\sigma_2 = 5.0$ and the wind velocity magnitude of $4~\si{m/s}$.

\begin{figure}[!t]
  \centering
  \includegraphics[width=\columnwidth, trim=2.7cm 2.5cm 1.2cm 3.2cm, clip]{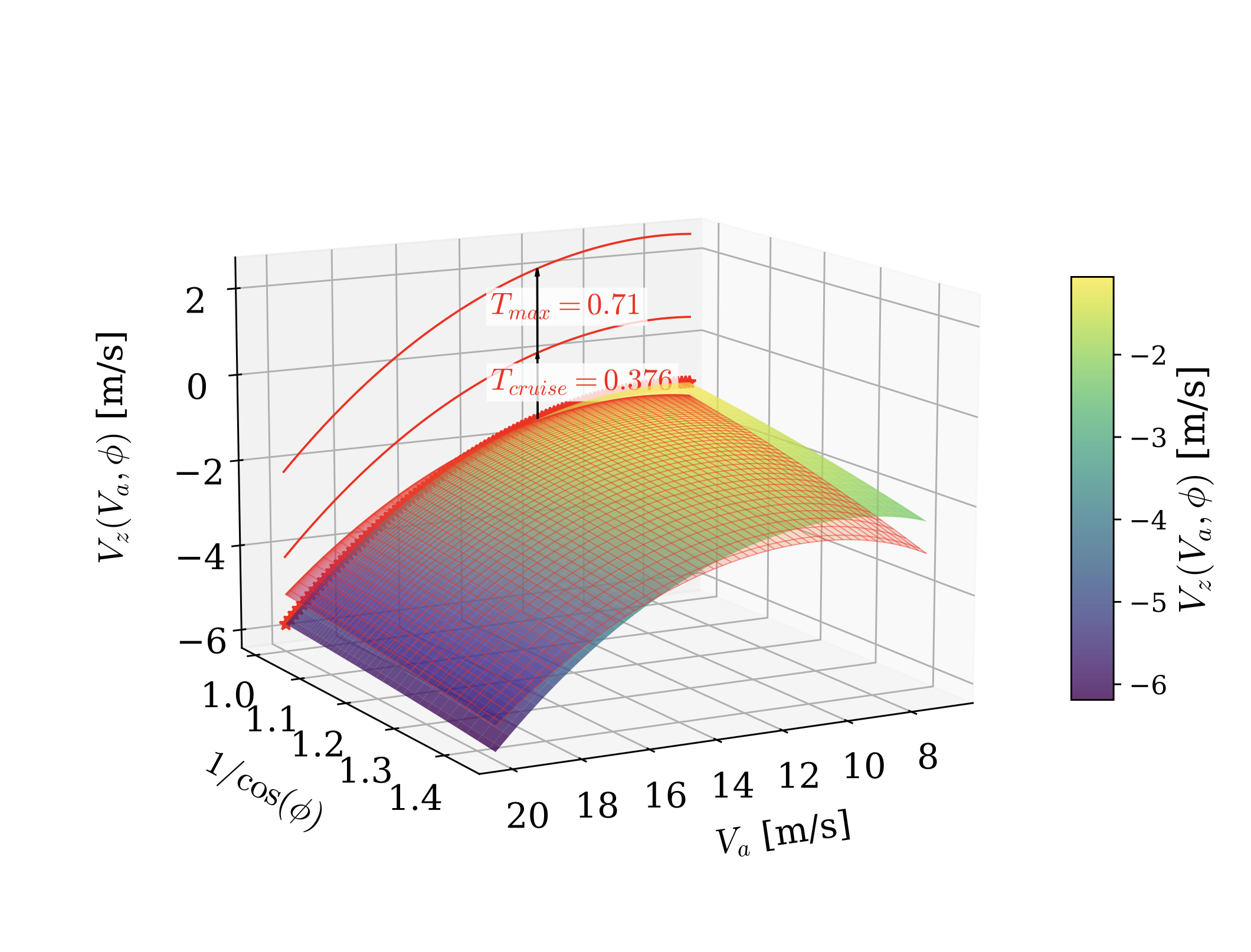}
  \caption{Sink Polar and relative sink rate as obtained through empirical data (green) and CFD analysis (red).
  \label{fig:sink_polar}}
  \vspace{-10pt}
\end{figure}

\subsection{Sink rate estimation}
To plan trajectories with respect to the aircraft sink rate $V_z(V_a, \phi)$, we estimate it using CFD simulations and real flight data. The resulting sink polars are shown in Fig.~\ref{fig:sink_polar}, with load $n = 1/cos \phi$ on the $y$-axis, and airspeed $V_a$ on the $x$-axis. Data were collected over  $20+$ glides
between $\mathbf{B}_{G, start}$ and $\mathbf{B}_{G, end}$ varying the slope angle $\gamma$ for each reference airspeed $V_{a, ref} \in [10, 18]~\si {m/s}$.
For each $V_{a, ref}$, the corresponding $V_z$ is extracted to construct the sink polar. The tangent from the origin identifies the airspeed that maximizes $L/D$, minimizing drag during the glide. Both CFD and experimental polars indicate an optimal gliding speed of $V_{a, ref}^* \approx~12~\si{m/s}$, yielding $V_z = 1.51~\si{m/s}$.
Given the $V_{a, ref}^*$, we further identify the proper amount of throttle required to cruise condition such that $V_z = 0$, defined as $u_{cruise} = 0.376 \in [0,1]$, as visible in red in Fig. \ref{fig:sink_polar}.

\subsection{Real world experiments}

\begin{figure}[!t]
  \centering
  \includegraphics[width=0.9\columnwidth,  trim=0.5cm 0.1cm 0.1cm 1.0cm, clip]{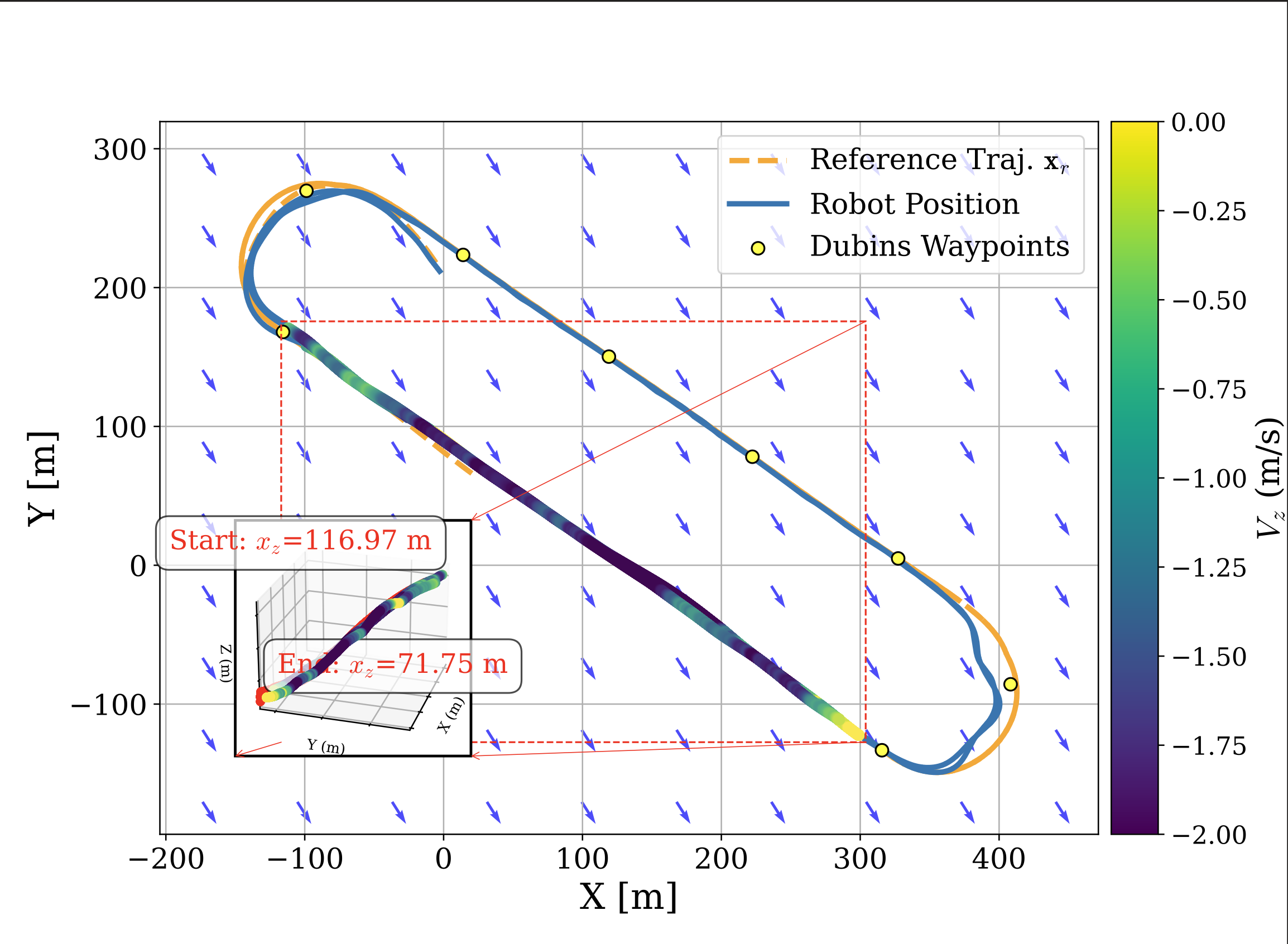}
  \caption{2D real world mission visualization in presence of $5~\si{m/s}$ tailwind. The recorded sink rate is stable at $V_z(V_a, \phi) =  -1.48~\si{m/s}$.
  \label{fig:2D_glide_no_obstacle}}
  \vspace{-20pt}
\end{figure}

\begin{figure}[!t]
  \centering
  \includegraphics[width=0.9\columnwidth,  trim=1.5cm 0.5cm 0.8cm 1.2cm, clip]{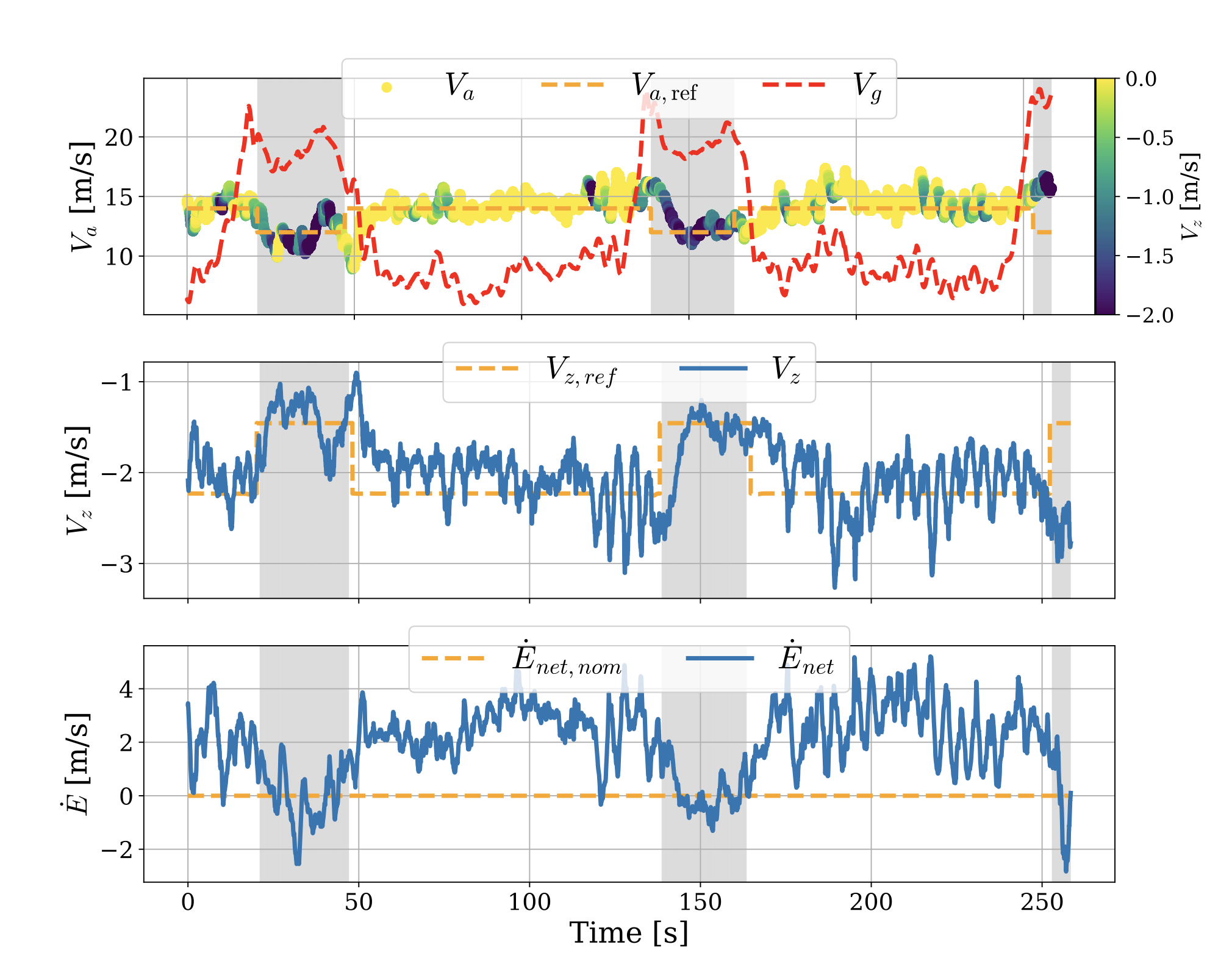}
  \caption{Glide Properties. The gray region corresponds to unactuated flight. The aircraft stabilizes at $V_{a,ref}$ with $\dot{E}_{net} \approx 0$, achieving steady glide despite wind, visible between the ground speed $V_g$ (red) and airspeed $V_a$ offset in the top plot.
\label{fig:airspeed_vario_no_obstacle}}
  \vspace{-20pt}
\end{figure}

\begin{figure}[!t]
  \centering
  \includegraphics[width=0.9\columnwidth,  trim=0.2cm 0.2cm 0.2cm 0.2cm, clip]{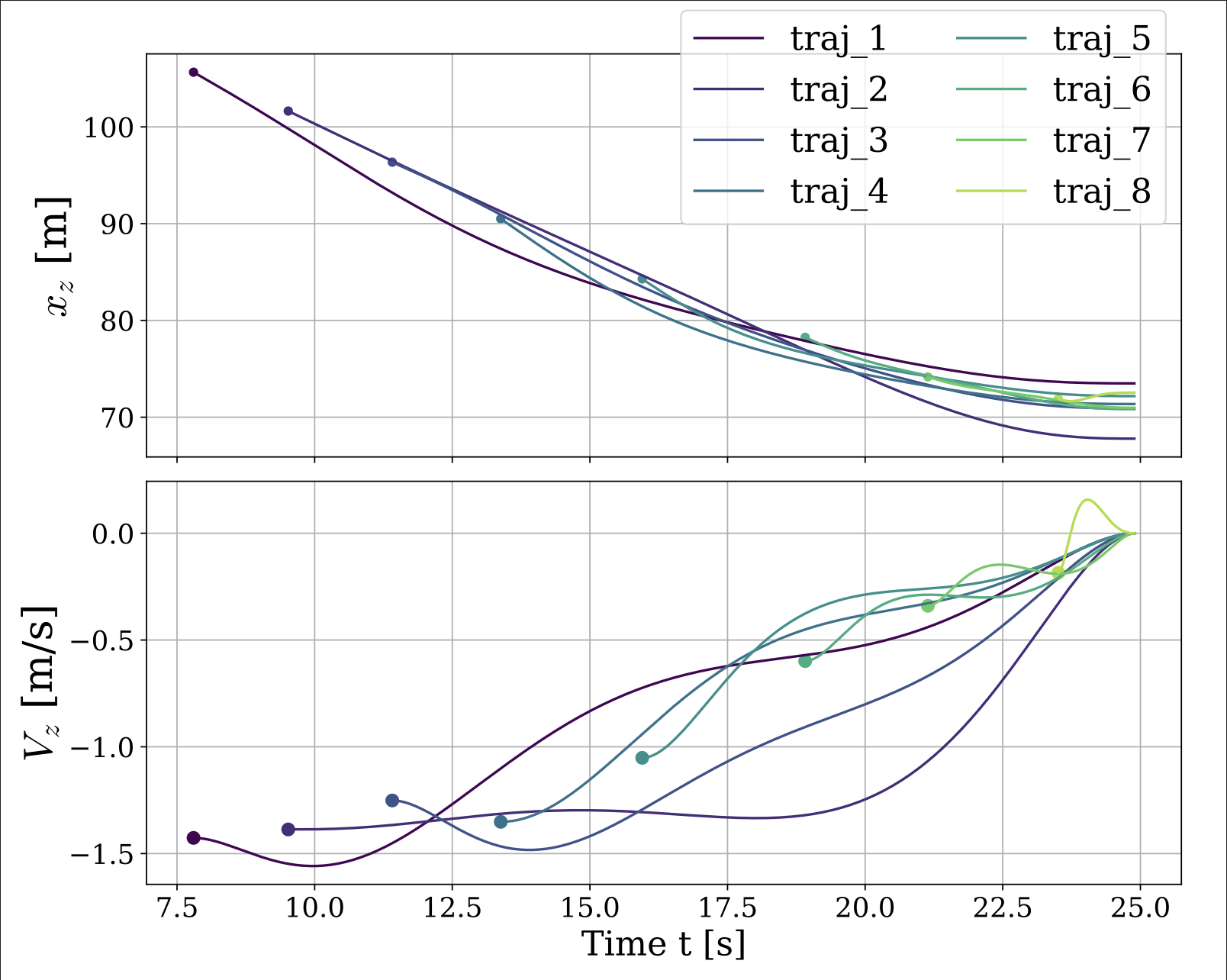}
  \caption{Continuous replanning adjusts the predicted sink rate based on most recent wind estimations.\label{fig:predicted_sink_rate_no_obstacle}}
  \vspace{-20pt}
\end{figure}

A sample real world mission is visible in 3D in Fig.~\ref{fig:fig1} and in 2D with wind directionality in Fig.~\ref{fig:2D_glide_no_obstacle}.
performing consecutive glides between $\mathbf{B}_{G,start}$ and $\mathbf{B}_{G,end}$, approximately $400~\si{m}$ apart, starting at $120~\si{m}$ altitude. The planner computes the altitude of the end waypoint and the slope of the trajectory and the vertical velocity based on the airspeed, sink rate and wind, while onboard sensors continuously estimate energy, airspeed and wind conditions. After each glide, a trajectory (sampled on 8 waypoints derived from a Dubins path) returns the aircraft to the starting waypoint. We selected $V_{a, ref} \approx 15~\si{m/s}$ during actuated flight, for higher wind resilience, while we decrease it to $V_{a, ref} \approx~12.5~\si{m/s} $ during the glide part, given an onboard estimated wind of $5.2~\si{m/s}$ in south west direction. 

During the colorful glide in Fig.~\ref{fig:2D_glide_no_obstacle}, the planner successfully stabilizes the aircraft to the nominal values of airspeed and sink rate, respectively $V_a = V_{a, ref} =12.5~\si{m/s}$ at $Vz = V_{z, ref} =-1.48~\si{m/s}$, such that $\dot{E}_{net} \approx 0$, as visible in Fig.~\ref{fig:airspeed_vario_no_obstacle}, throughout the entirety of the various glides, despite a recorded speed on the ground of $V_g \approx~20~\si{m/s}$, showing the important tailwind effect. 
Referring to the results shown in Fig.~\ref{fig:airspeed_vario_no_obstacle}, we demonstrate the accuracy of the planner in generating gliding trajectories such that the $RMSE_{V_a} = 1.09~\si{m/s}$, $RMSE_{V_z} =~0.39~\si{m/s}$ and $RMSE_{\dot{E}_{net}} =~0.69~\si{m/s}$ over a net glide length of $489.77~\si{m}$. 
From the recorded data, it is possible to quantify a net aircraft glide ratio of $ 9.83:1~\si{m}$, which is pretty low compared a standard sailplane glide ratio of $70:1~\si{m}$. However, the results shows effectivness of the planner and the re-planning technique shown in Fig.~\ref{fig:predicted_sink_rate_no_obstacle}, which immediately stabilizes the plane on a balance glide despite  the mechanical structural properties of the platform, that are not comparable to a glider. 

\subsection{Real world experiments with obstacle avoidance}
\begin{figure}[!t]
  \centering
  \includegraphics[width=\columnwidth,  trim=0.1cm 1.0cm 0.1cm 1.5cm, clip]{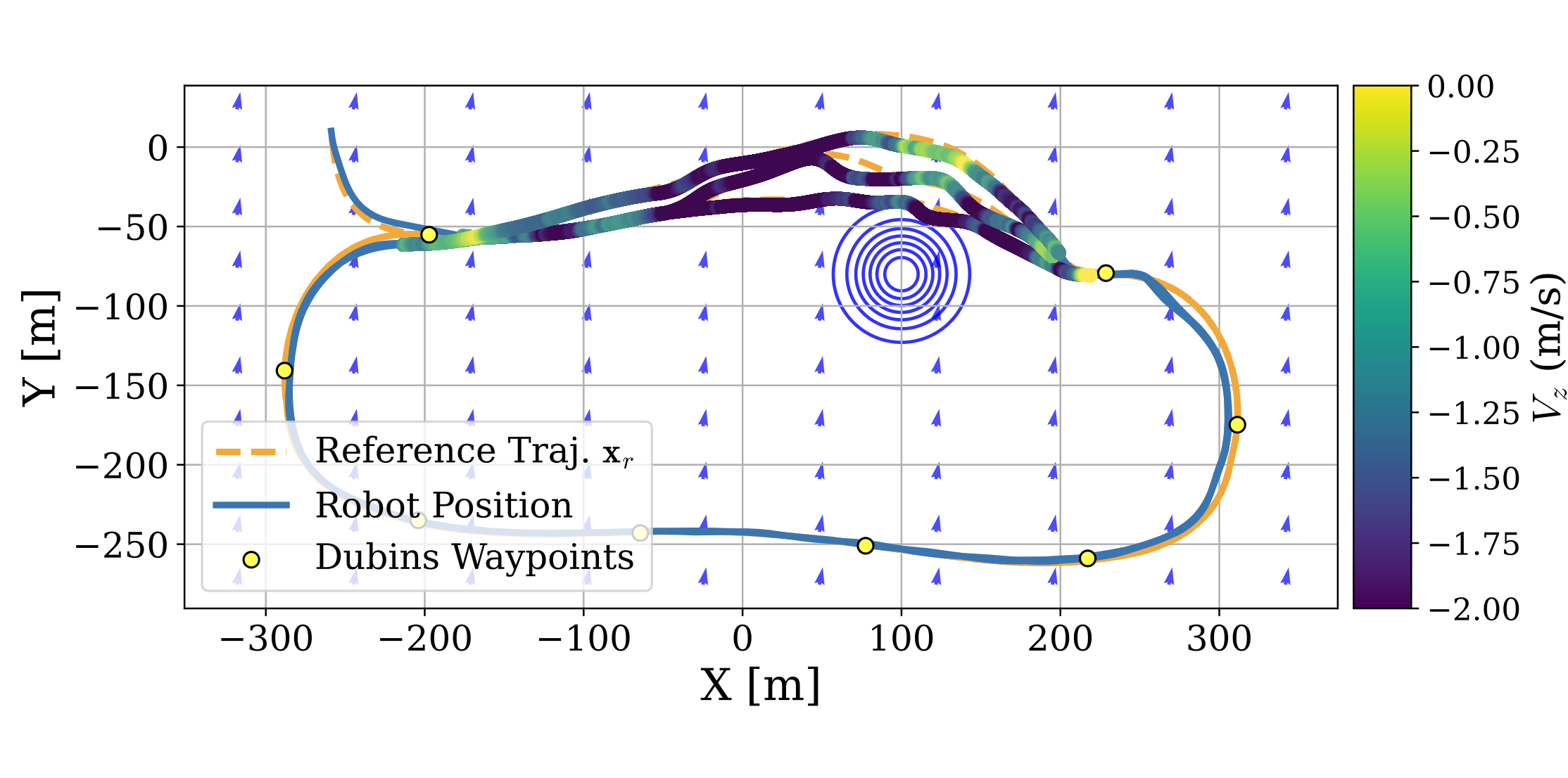}
  \caption{2D mission visualization in presence of gaussian obstacle and lateral estimated wind. 
  \label{fig:glide_2D_obstacle}}
  \vspace{-10pt}
\end{figure}

\begin{figure}[!t]
  \centering
  \includegraphics[width=0.9\columnwidth, trim=0.5cm 0.5cm 0.5cm 1.3cm, clip]{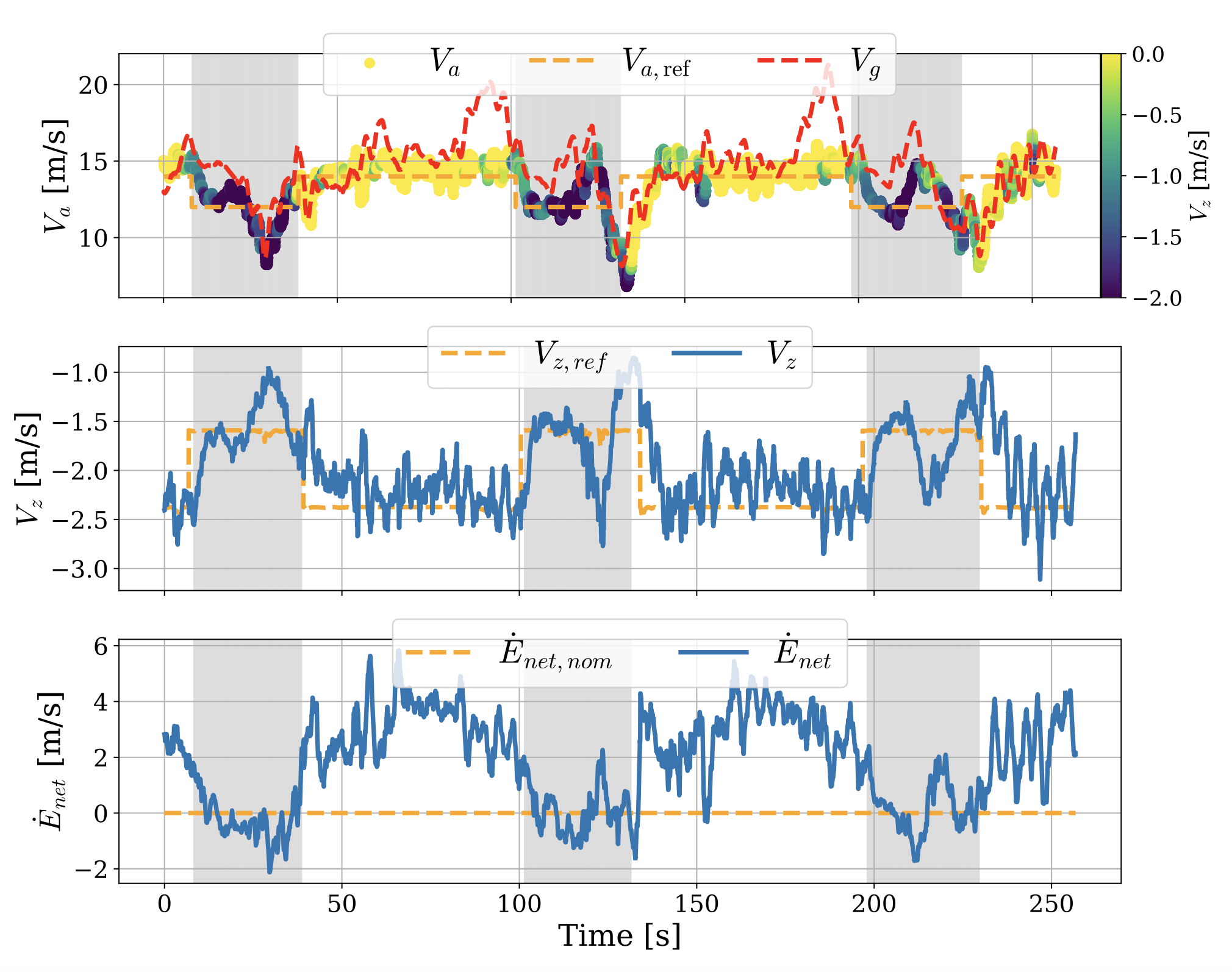}
  \caption{Gliding details based on recorded $V_a$, $V_z(V_a, \phi)$ and $\dot{E}_{net}$ in case of deviations due by the obstacle. 
  \label{fig:airspeed_vario_obstacle}}
  \vspace{-20pt}
\end{figure}


We also evaluate the planner in  challenging scenarios where an obstacle is present along the glide path.
A Gaussian obstacle with variance $\sigma_x = \sigma_y = 50~\si{m}$ and height $z = 90~\si{m}$ was placed at position $\mathbf{O}^I = [100, -80]$ $\si{m}$, as visible in Fig.~\ref{fig:glide_2D_obstacle}.
The reference gliding airspeed was set to $V_{a, ref} = 12.5~\si{m/s}$ as in the previous experiment. Even in this scenario, the planner successfully generated trajectories that quickly stabilized the aircraft at the desired gliding airspeed and nominal sink rate, as visible in Fig.~\ref{fig:airspeed_vario_obstacle}, while ensuring a minimum safe distance of $d_{safe} = 20~\si{m}$.
Some oscillations are observed when the trajectory crosses the wind direction at $\ang{90}$, a singular configuration where the wind cost becomes negligible.
In this case, the aircraft compensates by trading altitude for speed, which explains the temporary increase in airspeed visible in Fig.~\ref{fig:airspeed_vario_obstacle}. 
Despite this, the performance remained within acceptable limits, with RMSE values of $RMSE_{V_a} = 1.31~\si{m/s}$,  $RMSE_{V_z} = 0.42~\si{m/s}$ and $RMSE_{\dot{E}_{net}} = 0.78~\si{m/s}$. The glide covers a distance of $425.86~\si{m}$ with an altitude loss of $59.65~\si{m}$, showing a net glide ratio of $7.10:1~\si{m}$. This reduction compared to the previous test is expected and mainly due to the heading variation imposed by the obstacle and the additional energy required to counteract the wind.

\section{Conclusion}
\label{sec:Conclusion}

In this work, we presented a real-time planner for efficient gliding in windy conditions. Starting from steady-state glide analysis, we introduce an energy-based method that embeds netto variometer equations directly into the optimization.
The sink-rate polar is obtained from empirical and CFD data, and the approach is validated in both simulation and real-world tests. Using differential flatness, we generate fast, online Bernstein trajectories for gliding and cruising, yielding smooth paths that exploit favorable wind.

Experiments under varying winds and obstacles demonstrate the planner’s robustness in maintaining steady glide. Future work includes extending the method to soaring, where atmospheric lift supports sustained flight, and integrating the energy-based constraints directly into an NMPC formulation for short-horizon, glide-aware trajectories.



\bibliographystyle{IEEEtran}
\bibliography{reference}

\end{document}